\documentclass[runningheads]{llncs}

\usepackage[T1]{fontenc}
\usepackage{pifont}
\usepackage{amssymb}
\usepackage[table]{xcolor}
\usepackage{multirow}
\usepackage{booktabs}
\usepackage{graphicx,verbatim}
\usepackage{amsfonts}
\usepackage{amsmath} 
\usepackage{graphicx}
\usepackage[misc]{ifsym}

\newcommand{\cmark}{\ding{51}}%
\newcommand{\xmark}{\ding{55}}%
\newcommand{\corrauth}{\fnmsep\textsuperscript{\scalebox{0.75}{\Letter}}}

\begin{document}
\title{Latent-CURE: Interpretable Breast Cancer Diagnosis via Dual-Asymmetric Chain-of-Thought}
%
\titlerunning{Latent-CURE for Breast Cancer Diagnosis}

\author{
Weiyi Zhao\inst{1} \and
Xiaoyu Tan\inst{2} \and
Lu Gan\inst{3} \and
Liang Liu\inst{3} \and
Xihe Qiu\inst{1}\corrauth
}


\authorrunning{W. Zhao et al.}

\institute{
School of Electronic and Electrical Engineering, Shanghai University of Engineering Science, Shanghai 201620, China\\
\email{qiuxihe1993@gmail.com}
\and
Tencent YouTu Lab, Shanghai 200232, China
\and
Integrative Clinical Research Ward, Clinical Medicine Research Institute, Zhongshan Hospital, Fudan University, Shanghai 200032, China
}

\maketitle
\begingroup
\renewcommand{\thefootnote}{}
\footnotetext{\Letter~Corresponding author.}
\endgroup

\begin{abstract}
Multimodal Large Models have significantly advanced automated breast ultrasound diagnosis. However, most existing frameworks utilize opaque, end-to-end paradigms prioritizing global statistical correlations over structured clinical reasoning. Consequently, these models remain susceptible to shortcut learning amid extreme real-world epidemiological imbalances, often bypassing rare but decisive malignant indicators for dominant benign patterns. To address this disconnect, we propose Latent-CURE, a novel diagnostic framework driven by asymmetric weighted chain-of-thought methodology grounded in latent space reasoning. Unlike traditional approaches, our framework constructs an implicit reasoning trajectory forcing the model to sequentially infer standardized BI-RADS morphological descriptors before converging on a final diagnosis. Furthermore, to combat the extreme scarcity of critical malignant features, we couple this architecture with a dual-asymmetric optimization strategy. By dynamically adjusting margins and weights, this strategy safeguards high-specificity malignant descriptors from being overshadowed by common benign priors. Comprehensive evaluations demonstrate that our knowledge-injected approach provides transparent clinical evidence while achieving robust, accurate diagnostic performance in imbalanced medical cohorts.

\end{abstract}
\section{Introduction}

Breast cancer detection relies heavily on breast ultrasound (BUS), a non-invasive modality requiring radiologists to meticulously evaluate morphological features, including margins, echogenicity, and orientation, according to the standardized Breast Imaging-Reporting and Data System (BI-RADS) \cite{1brem2015screening,1xian2018automatic,1xu2019medical}. To reduce diagnostic variability and operator dependence, clinical artificial intelligence has rapidly transitioned toward the adoption of Multimodal Large Models (MLMs) \cite{16singhal2023large,18thirunavukarasu2023large,19amann2020explainability}. Pre-trained on massive datasets, these foundational architectures exhibit remarkable representational capabilities, driving state-of-the-art performance in automated lesion classification and segmentation \cite{17li2023llava,20hao2025large}.

\begin{figure}[ht]
 \centering
 \includegraphics[width=0.9\linewidth]{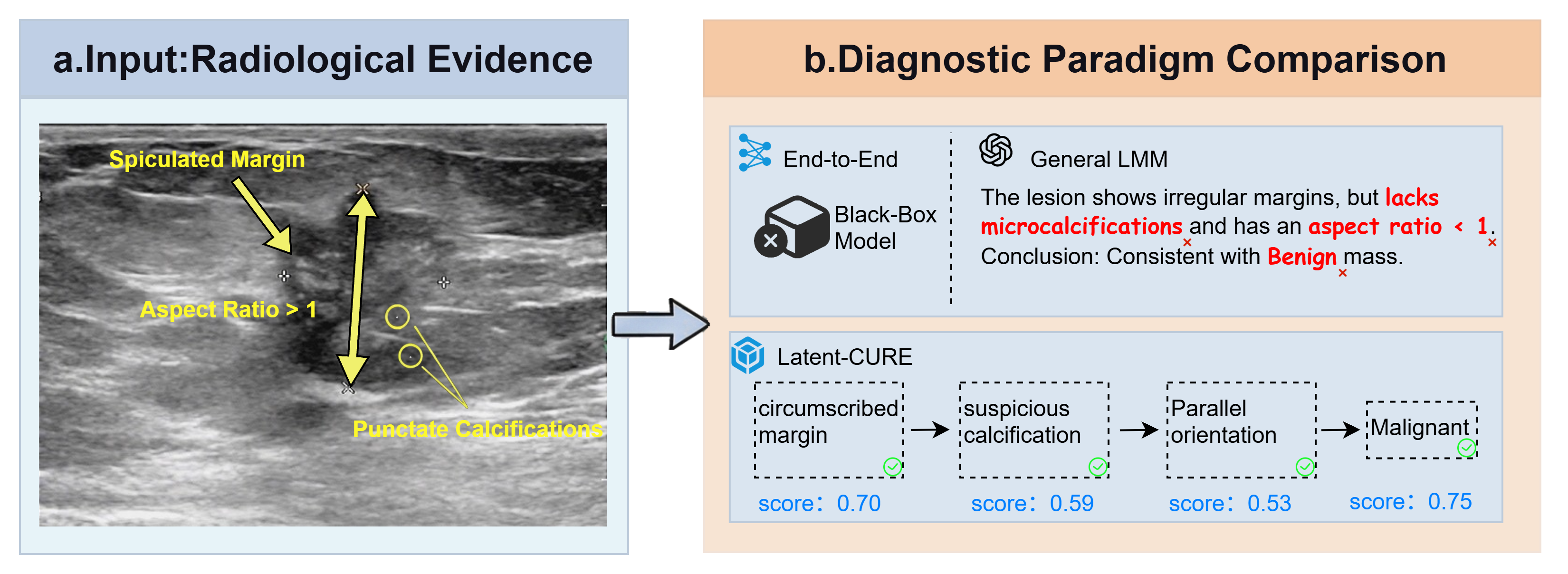}
 \caption{Diagnostic challenge and our proposed paradigm. Compared to black-box LMMs prone to shortcut learning, Latent-CURE ensures transparent diagnosis via step-wise morphological reasoning.}
 \label{fig:motivation}
\end{figure}

Despite the representational power of these systems, existing LMM-based diagnostic frameworks predominantly operate under an opaque end-to-end paradigm that maps deep visual embeddings directly to diagnostic labels, prioritizing global statistical correlations over structured clinical reasoning \cite{21zheng2025review,22qian2025multimodal,COT}. This creates a critical vulnerability rooted in the extreme epidemiological imbalance of real-world medical cohorts, where benign lesions vastly outnumber malignant cases \cite{1saleem2025hybrid}. Consequently, even advanced foundation models are highly susceptible to shortcut learning. Rather than identifying statistically rare but decisive malignant indicators, such as non-parallel growth or subtle microcalcifications, these models tend to bypass them in favor of dominant, easily recognizable benign patterns, as illustrated in Fig.~\ref{fig:motivation}. This failure to navigate clinical imbalance fundamentally severs the link between the latent decisions of the model and rigorous expert logic \cite{23kolla2024uses,24holmes2023evaluating,26choi2024availability}.

\begin{figure}[!t]
 \centering
 \includegraphics[width=0.9\linewidth]{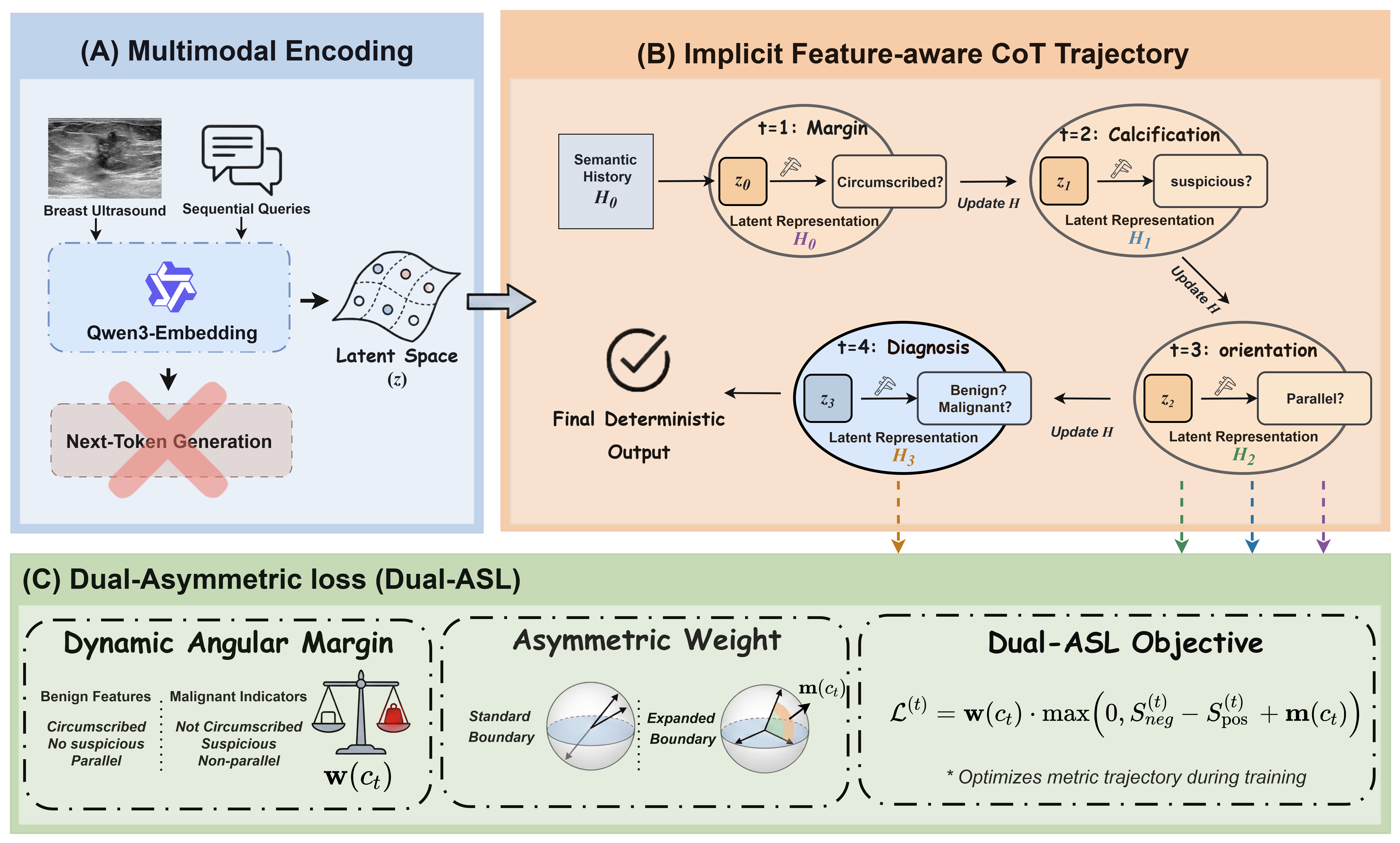}
 \caption{Overview of the proposed Latent-CURE architecture. The framework consists of multimodal encoding, an implicit feature-aware CoT trajectory, and a Dual-ASL strategy.}
 \label{fig:architecture}
\end{figure}

To address this disconnect, we propose an asymmetric weighted chain-of-thought methodology grounded in latent space reasoning (illustrated in Fig.~\ref{fig:architecture}). Unlike traditional end-to-end methods, our framework exploits the autoregressive properties of advanced embedding spaces to construct an implicit reasoning trajectory \cite{2hao2024training,2shen2025codi,2su2025token,2tan2025think}, forcing the model to sequentially infer key clinical BI-RADS descriptors before converging on a final diagnosis. To combat the extreme scarcity of critical malignant features, we couple this architecture with a novel dual-asymmetric loss (Dual-ASL) function. This strategy dynamically adjusts learning priorities, safeguarding high-specificity malignant descriptors from being overshadowed by common benign patterns to ensure transparent and robust diagnostic performance.

\noindent \textbf{Contributions.} The primary contributions are three-fold: 
(1) \textbf{Knowledge-Injected Latent Reasoning:} an implicit Chain-of-Thought (CoT) approach embedding expert descriptors into the latent process; 
(2) \textbf{Explainability via Professional Descriptors:} a transparent alternative to end-to-end models with evidence-supported diagnosis; 
(3) \textbf{Dual-Asymmetric Optimization:} a novel loss strategy using dynamic margins to protect rare, high-specificity malignant features.

\section{Methodology}

\subsection{Problem Formulation}
\label{sec:formulation}

In the context of MLMs, the conventional automated diagnostic process is typically formulated as an autoregressive generation task. Given a BUS image $\mathbf{X}_{v} \in \mathbb{R}^{H \times W \times C}$ and a textual prompt $\mathbf{X}_{p}$, traditional MLMs generate a sequence of textual tokens $\mathbf{Y} = (y_1, y_2, \dots, y_N)$ by maximizing the conditional probability $P(\mathbf{Y} | \mathbf{X}_{v}, \mathbf{X}_{p}) = \prod_{i=1}^{N} p(y_i | \mathbf{X}_{v}, \mathbf{X}_{p}, y_{<i})$. Although versatile, this unconstrained next-token prediction paradigm inevitably introduces semantic hallucinations \cite{2zhang2025soft,2zhu2025surveylatentreasoning}. This vulnerability renders the approach unsafe for rigorous medical decision-making, a domain that strictly requires deterministic outputs.

To eliminate hallucinations at the architectural level, we reformulate the multimodal diagnostic process from an open-ended generation task to a deterministic metric learning problem within a normalized latent space. Let $f_\theta(\cdot)$ denote the unified multimodal encoder parameterized by $\theta$, which is capable of projecting multimodal contexts and unimodal text into a shared $d$-dimensional hypersphere.  For a specific diagnostic sub-task $t$, we define a set of mutually exclusive clinical semantic anchors $\mathcal{A}_t = \{a_{pos}^{(t)}, a_{neg}^{(t)}\}$. The objective is to retrieve the target anchor that maximizes the cosine similarity with the given multimodal context:
\begin{equation}
 \hat{a}^{(t)} = \arg\max_{a \in \mathcal{A}_t} \cos \left( f_\theta(\mathbf{X}_{v}, \mathbf{X}_{p}), f_\theta(a) \right)
\end{equation}

By replacing the probabilistic decoding head with a deterministic retrieval-based objective, the output space is strictly bounded by predefined expert knowledge, which theoretically guarantees a zero-hallucination diagnostic trajectory.

\subsection{Implicit Feature-aware Chain-of-Thought Modeling}
\label{sec:cot}

Although the metric learning formulation presented in Sec.~\ref{sec:formulation} guarantees deterministic outputs, a direct mapping from the image $\mathbf{X}_v$ to the final diagnosis bypasses the intermediate morphological reasoning required by the BI-RADS standard. To bridge this gap, we propose an implicit feature-aware CoT module, which forces the latent representations to align sequentially with predefined clinical descriptors prior to the synthesis of the final decision.

We model the diagnostic process as a discrete Markovian decision sequence of length $T$. Let $\mathcal{T} = \{1, 2, \dots, T\}$ denote the set of ordered reasoning steps. Based on established radiological cognitive pathways, we set $T=4$. This sequence reflects a diagnostic progression from macro-structures to micro-signals, culminating in a holistic assessment: Margin ($t=1$) $\to$ Calcification ($t=2$) $\to$ Orientation ($t=3$) $\to$ Final Diagnosis ($t=4$). 

At each step $t \in \mathcal{T}$, the model resolves a specific clinical query $\mathbf{q}_t$. To preserve the autoregressive nature of the reasoning process without requiring explicit text generation, the input at step $t$ is conditioned on the visual context $\mathbf{X}_v$ and the accumulated semantic history $\mathbf{H}_{t-1}$ from all preceding steps. This history is constructed by concatenating deterministically selected anchors:
\begin{equation}
\mathbf{H}_{t-1} = \bigoplus_{i=1}^{t-1} \hat{a}^{(i)}, \quad \text{with} \quad \mathbf{H}_0 = \emptyset
\end{equation}
where $\oplus$ denotes textual sequence concatenation.

The context-aware latent representation $\mathbf{z}_t \in \mathbb{R}^d$ for the $t$-th reasoning step is extracted by the multimodal representation encoder $f_\theta$ as follows:
\begin{equation}
\mathbf{z}_t = f_\theta \left( \mathbf{X}_v \oplus \mathbf{q}_t \oplus \mathbf{H}_{t-1} \right)
\end{equation}

The optimal diagnostic anchor at step $t$ is determined through nearest neighbor retrieval in the latent space:
\begin{equation}
\hat{a}^{(t)} = \arg\max_{a \in \mathcal{A}_t} \cos(\mathbf{z}_t, f_\theta(a))
\end{equation}

This sequential formulation ensures that the embedding vector $\mathbf{z}_4$ utilized for the final classification of malignancy is intrinsically conditioned on explicit anatomical findings. Embedding this causal chain directly into the latent space achieves a highly interpretable diagnostic trajectory wherein the ultimate prediction remains deterministically anchored to explicit, intermediate BI-RADS evidence.

\subsection{Dual-Asymmetric Optimization for Long-tailed Clinical Features}
\label{sec:asl}

Although the implicit CoT reasoning integrates clinical pathways, optimizing this trajectory under real-world epidemiological distributions presents fundamental challenges. Medical datasets are notoriously long-tailed; benign lesions consistently outnumber malignant ones, and specific malignant morphological indicators, such as non-parallel orientation or microcalcifications, remain statistically scarce. Under standard symmetric optimization objectives, the gradient landscape is overwhelmingly dominated by the abundant benign visual priors \cite{ASL}. This imbalance invariably causes gradient starvation for rare features, which collapses the latent space toward majority classes and exacerbates the shortcut learning phenomenon.  To mitigate this majority collapse, we propose a Dual-ASL tailored for the metric-based reasoning steps. This strategy injects asymmetry into two orthogonal dimensions: margin adaptation for boundary strictness and weight penalty for gradient magnitude, ensuring that scarce but critical malignant anchors dictate the optimization trajectory.
For a given reasoning step $t \in \mathcal{T}$ with the extracted latent state $\mathbf{z}_t$, let $S_{pos}^{(t)} = \cos(\mathbf{z}_t, f_\theta(a_{pos}^{(t)}))$ and $S_{neg}^{(t)} = \cos(\mathbf{z}_t, f_\theta(a_{neg}^{(t)}))$ denote the cosine similarities between the visual embedding and the corresponding textual anchors. To prevent majority collapse, the penalty is conditioned not solely on the step $t$, but strictly on the specific ground-truth clinical feature $c_t$. We formulate the step-wise Dual-ASL objective as a dynamically weighted cosine triplet margin loss:
\begin{equation}
 \mathcal{L}_{\text{Dual-ASL}}^{(t)} = \mathbf{w}(c_t) \cdot \max \left(0, S_{neg}^{(t)} - S_{pos}^{(t)} + \mathbf{m}(c_t) \right)
\end{equation}
where $\mathbf{w}(c_t)$ acts as the feature-specific dynamic sample weight and $\mathbf{m}(c_t)$ defines the adaptive angular margin.

The values of $\mathbf{w}(c_t)$ and $\mathbf{m}(c_t)$ are dynamically conditioned on the epidemiological prior and the diagnostic specificity of the target anchor $a_{pos}^{(t)}$. The margin $\mathbf{m}(c_t)$ forces a stricter latent separation boundary for malignant-associated features, compelling the encoder $f_\theta$ to learn a discriminative manifold that resists benign feature entanglement. Simultaneously, the weight $\mathbf{w}(c_t)$ aggressively up-weights structurally rare anchors. By amplifying the gradient signals of these high-specificity, low-frequency indicators, the proposed formulation effectively shields them from being overshadowed. 

The global optimization objective is the summation of the asymmetric losses across all implicit reasoning steps:
\begin{equation}
 \mathcal{L}_{total} = \sum_{t=1}^{T} \mathcal{L}_{\text{Dual-ASL}}^{(t)}
\end{equation}
This architecture ensures that the model cannot achieve a low global loss by merely defaulting to the prediction of benign outcomes; it must actively align with the rare pathological clues.

\section{Experiments}

\subsection{Experimental Setup}
\label{sec:experimental_setup}

We utilized an IRB-approved multicenter dataset of 666 breast ultrasound cases, partitioned 8:2 for training and testing, with all diagnostic labels strictly derived from the surgical histopathology gold standard. To eliminate irrelevant background interference and prevent shortcut learning, physicians manually cropped the images to isolate regions of interest. Following consultations with multidisciplinary physicians, we distilled BI-RADS descriptors into three core dimensions: Margin, Calcification, and Orientation. To combat epidemiological imbalance and prevent gradient starvation on rare malignant features, the model was optimized via our Dual-ASL framework. Comprehensive implementation details are summarized in Table \ref{tab:implementation_details}.

\subsection{Quantitative Baselines and Efficiency Analysis}

\begin{figure}[!t]
 \centering
 \includegraphics[width=0.9\linewidth]{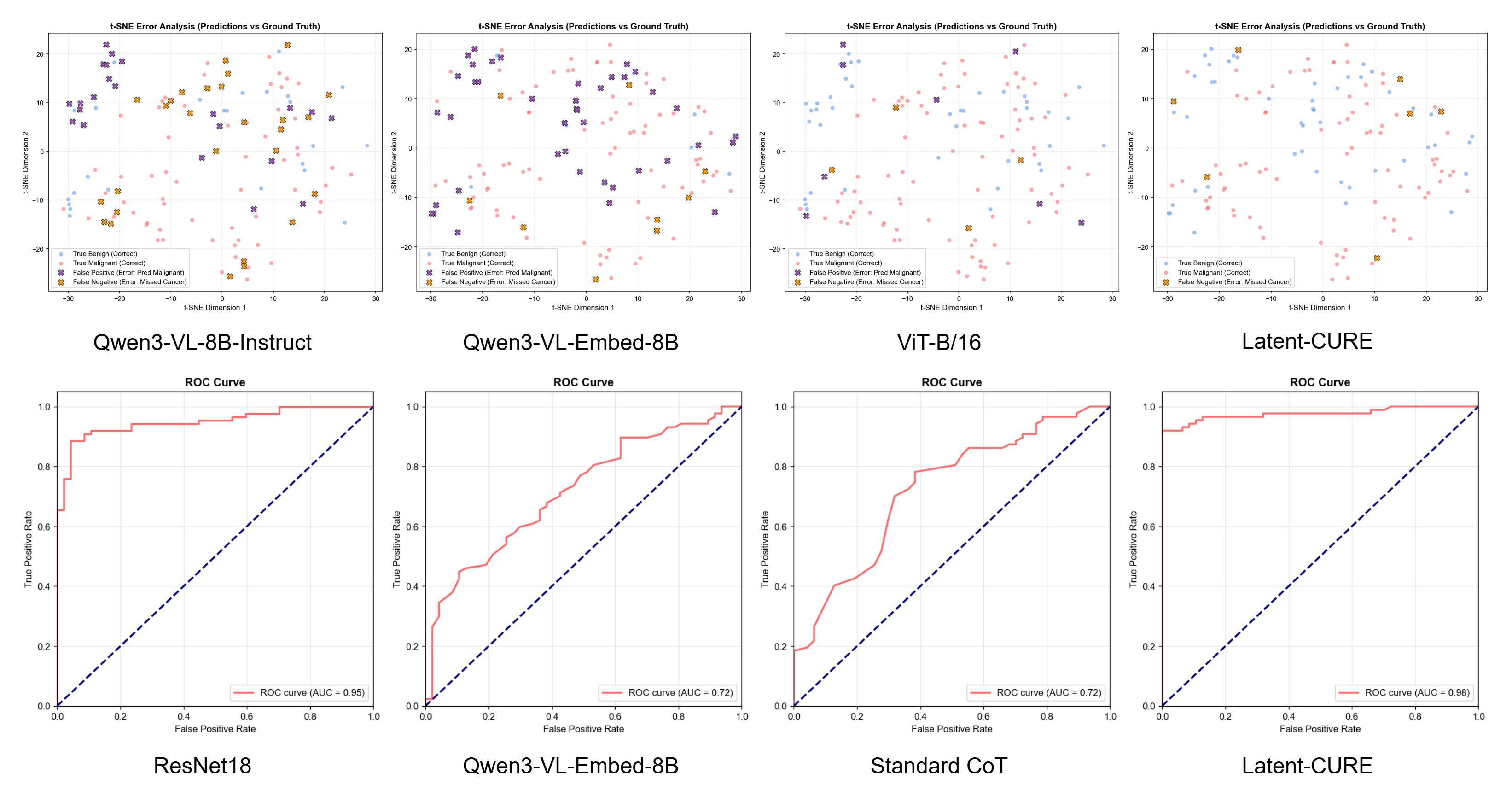}
 \caption{Qualitative and quantitative analysis of diagnostic models.}
 \label{fig:tsne_roc}
\end{figure}

\begin{figure}[!t]
 \centering
 \includegraphics[width=0.9\linewidth]{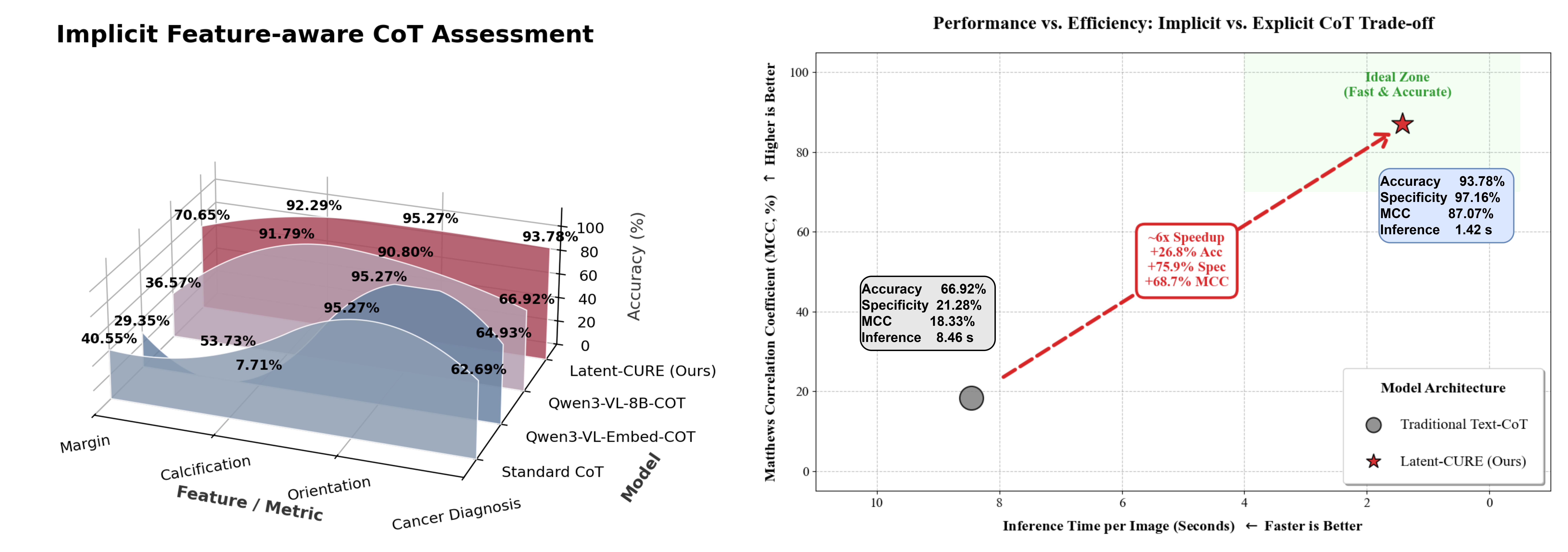}
 \caption{Implicit CoT Assessment: Accuracy Breakdown and Efficiency Trade-off}
 \label{fig:tradeoff}
\end{figure}

As summarized in Table~\ref{tab:main_results} and visualized in Fig.~\ref{fig:tsne_roc} and Fig.~\ref{fig:tradeoff}, Latent-CURE establishes a new state of the art, achieving 93.78\% accuracy and an 87.07\% Matthews correlation coefficient (MCC), systematically outperforming pure vision baselines such as ViT-B/16 and ResNet18. The evaluation simultaneously exposes severe limitations across general foundation models. Both large-scale proprietary frameworks and open adapted variants demonstrate a high susceptibility to shortcut learning and autoregressive hallucinations, overwhelmingly defaulting to the prediction of benign outcomes and causing significant specificity degradation that ranges from 21.28\% to 77.04\%. Latent-CURE circumvents these generative bottlenecks by executing the reasoning trajectory within a protected latent space guided by Dual-ASL.  Explicitly aligning this trajectory with clinical morphological descriptors effectively mitigates false positive susceptibility, which restores diagnostic precision to 97.16\% specificity while achieving an approximately sixfold inference acceleration over explicit text-based adaptation strategies.

\begin{table}[t]
\centering
\caption{Comprehensive Implementation Details: Global Hyperparameters and Feature-Specific Dual-ASL Configurations.}
\label{tab:implementation_details}

\fontsize{8pt}{9.6pt}\selectfont
\setlength{\tabcolsep}{2pt} 

\begin{tabular}{ll|ll}
\toprule
\multicolumn{4}{c}{\textbf{Part I: Global Training Hyperparameters}} \\
\midrule
\textbf{Parameter} & \textbf{Value} & \textbf{Parameter} & \textbf{Value} \\
\midrule
Hardware & GPU (80GB) & Base LR & $1 \times 10^{-5}$ \\
Base Model & Qwen3-VL-Embed-8B & LR Scheduler & Cosine \\
LoRA Config & $r=16, \alpha=32$ & Seeds & 42, 666, 2026 \\
\midrule
\midrule
\multicolumn{4}{c}{\textbf{Part II: Dual-ASL Configuration Matrix}} \\
\midrule
\textbf{Reasoning Step ($t$)} & \textbf{Ground-Truth ($c_t$)} & \textbf{Weight $\mathbf{w}$} & \textbf{Margin $\mathbf{m}$} \\
\midrule
\multirow{2}{*}{$t=1$: Margin} & Circumscribed (Benign) & 2.5 & 0.3 \\
 & Not Circumscribed (Malignant) & 1.0 & 0.2 \\
\midrule
\multirow{2}{*}{$t=2$: calcification} & No suspicious (Majority) & 1.0 & 0.2 \\
 & Suspicious (Malignant) & 2.0 & 0.3 \\
\midrule
\multirow{2}{*}{$t=3$: orientation} & Parallel (Majority) & 1.0 & 0.2 \\
 & Non-parallel (Rare) & \textbf{4.0} & \textbf{0.4} \\
\midrule
\multirow{2}{*}{$t=4$: Final} & Benign (Majority) & 1.0 & 0.2 \\
 & \textbf{Malignant (Target)} & \textbf{6.0} & \textbf{0.5} \\
\bottomrule
\end{tabular}
\end{table}

\begin{table}[!t]
\centering
\caption{Quantitative diagnostic performance comparison across different model architectures. Metrics are reported with standard deviations.}
\label{tab:main_results}

\fontsize{8pt}{9.6pt}\selectfont

\begin{tabular}{l|c|c|c|c}
\toprule
\textbf{Model Architecture} & \textbf{Accuracy (\%)} & \textbf{Precision (\%)} & \textbf{Specificity (\%)} & \textbf{MCC (\%)} \\
\midrule

ResNet18 \cite{resnet18} & 89.80 $\pm$ 2.70 & 97.01 $\pm$ 0.60 & 95.04 $\pm$ 1.10 & 79.50 $\pm$ 4.64 \\
ViT-B/16 \cite{ViT-B} & 92.29 $\pm$ 1.02 & 94.69 $\pm$ 2.88 & 90.07 $\pm$ 5.81 & 83.27 $\pm$ 2.61 \\
\midrule

Gemini-2.5-Pro \cite{gemini} & 73.13 $\pm$ 6.46 & 73.88 $\pm$ 3.45 & 40.43 $\pm$ 11.26 & 38.52 $\pm$ 17.57 \\
Qwen3-VL-Plus \cite{qwen3} & 77.36 $\pm$ 1.55 & 76.40 $\pm$ 0.63 & 46.10 $\pm$ 2.46 & 48.47 $\pm$ 4.57 \\
GPT-5 \cite{singh2025openai} & 78.11 $\pm$ 1.88 & 75.88 $\pm$ 2.23 & 42.55 $\pm$ 7.67 & 51.16 $\pm$ 4.29 \\
Baichuan-M2-Plus \cite{BAICHUAN} & 88.12 $\pm$ 10.51 & 89.44 $\pm$ 11.36 & 77.04 $\pm$ 23.34 & 73.33 $\pm$ 23.27 \\
\midrule

Qwen3-VL-Embed-8B \cite{qwen3embedding} & 58.71 $\pm$ 4.56 & 63.82 $\pm$ 2.42 & 12.06 $\pm$ 5.35 & -5.13 $\pm$ 11.98 \\
Qwen3-VL-Embed-COT & 64.93 $\pm$ 0.00 & 64.93 $\pm$ 0.00 & 0.00 $\pm$ 0.00 & 0.00 $\pm$ 0.00 \\
Qwen3-VL-8B-Instruct & 64.43 $\pm$ 0.77 & 71.91 $\pm$ 0.98 & 46.10 $\pm$ 5.49 & 20.83 $\pm$ 0.67 \\
Qwen3-VL-8B-lora \cite{lora} & 63.43 $\pm$ 2.00 & 73.22 $\pm$ 2.13 & 53.19 $\pm$ 5.03 & 21.70 $\pm$ 4.80 \\
Qwen3-VL-8B-COT & 66.92 $\pm$ 0.77 & 68.28 $\pm$ 0.28 & 21.28 $\pm$ 0.00 & 18.33 $\pm$ 2.20 \\
\midrule

\textbf{Latent-CURE (Ours)} & \textbf{93.78 $\pm$ 0.77} & \textbf{98.37 $\pm$ 1.26} & \textbf{97.16 $\pm$ 2.20} & \textbf{87.07 $\pm$ 1.85} \\

\bottomrule
\end{tabular}
\end{table}

\subsection{Ablation Studies}
\label{sec:ablation}

To evaluate individual module contributions, we conducted a systematic ablation study (Table~\ref{tab:ablation}). While the Direct SFT baseline provides a strong performance foundation (91.79\% accuracy, 96.45\% specificity), it remains an opaque classifier. Transitioning to a Standard CoT variant without targeted optimization triggers a catastrophic diagnostic collapse, where accuracy drops to 62.69\% and specificity plummets to 4.26\% with a negative -3.95\% MCC. This reveals the inherent instability of unconstrained reasoning in large multimodal models, where generating morphological anchors without gradient protection exacerbates majority class bias and shortcut learning. Integrating both Implicit CoT and Dual-ASL, the Latent-CURE framework effectively mitigates this collapse and surpasses the baseline across all metrics. By shielding scarce malignant anchors from gradient starvation, Dual-ASL elevates accuracy to 93.78\% and specificity to 97.16\%, with MCC improving from 83.26\% to 87.07\%. These gains validate that explicit gradient protection is mandatory to transform clinical reasoning into a deterministic, highly discriminative diagnostic trajectory.

\begin{table}[!t]
\centering
\caption{Ablation studies on the core components of Latent-CURE. \cmark~indicates the inclusion of a module. Metrics are reported as percentages (\%).}
\label{tab:ablation}

\fontsize{8pt}{9.6pt}\selectfont

\begin{tabular}{l|cc|c|c|c}
\toprule
\textbf{Model Variant} & \textbf{Implicit CoT} & \textbf{Dual-ASL} & \textbf{Accuracy} & \textbf{Specificity} & \textbf{MCC} \\
\midrule
Direct SFT & \xmark & \xmark & 91.79 $\pm$ 0.75 & 96.45 $\pm$ 1.23 & 83.26 $\pm$ 1.40 \\
Standard CoT & \cmark & \xmark & 62.69 $\pm$ 1.77 & 4.26 $\pm$ 3.30 & -3.95 $\pm$ 6.12 \\
\midrule
\textbf{Latent-CURE } & \cmark & \cmark & \textbf{93.78 $\pm$ 0.77} & \textbf{97.16 $\pm$ 2.20} & \textbf{87.07 $\pm$ 1.85} \\
\bottomrule
\end{tabular}
\end{table}

\section{Discussion and Conclusion}
We present Latent-CURE, a framework bridging multimodal models and clinical logic by shifting breast ultrasound diagnosis from open-ended generation to an implicit reasoning trajectory within a constrained latent space. Rather than relying on free-form CoT decoding, Latent-CURE performs deterministic retrieval over fixed BI-RADS descriptor anchors, thereby providing bounded intermediate evidence before the final diagnosis. While our dual-asymmetric optimization helps protect rare malignant indicators from dominant benign priors, the current reasoning trajectory follows a fixed descriptor order from margin, calcification, and orientation to the final diagnosis, and is not assumed to be permutation-invariant. Future work should further study how alternative descriptor sequences and feature-specific contributions affect latent representations and diagnostic outcomes. Our conclusions are limited to the submitted breast ultrasound setting with histopathology-derived labels. Overall, Latent-CURE demonstrates that standardized latent-anchor reasoning combined with asymmetric optimization can improve both interpretability and robustness in imbalanced clinical diagnosis.

\begin{credits}

\subsubsection{\discintname}
The authors have no competing interests to declare that are relevant to the content of this article.
\end{credits}

%
%
%
\bibliographystyle{splncs04}
\bibliography{ref}

\end{document}